\documentclass[letterpaper, 10 pt, conference]{ieeeconf}  

\usepackage{amssymb}  %
\usepackage{amsmath}
\usepackage{graphicx}
\usepackage{cite}
\usepackage{xcolor} %
\usepackage{colortbl,booktabs} %
\usepackage{multirow} %
\usepackage{color} %
\usepackage{hyperref}
\usepackage{tikz}
\definecolor{deeppeach}{rgb}{1.0, 0.8, 0.64}  %
\definecolor{lightgray}{rgb}{0.83, 0.83, 0.83}  %

\IEEEoverridecommandlockouts                              

\title{\LARGE \bf
ADGaussian: Generalizable Gaussian Splatting for Autonomous Driving via Multi-modal Joint Learning
}

\author{Qi Song$^{1}$, Chenghong Li$^{1}$, Haotong Lin$^{2}$, Sida Peng$^{2}$, Rui Huang$^{1\dag}$
\thanks{*This work was supported by Shenzhen Science and Technology Program under grant No. JCYJ20220818103006012 and ZDCY20250901103359008.}
\thanks{$^{1}$School of Science and Engineering, The Chinese University of Hong Kong (Shenzhen), Longgang, Shenzhen, Guangdong, 518172, P.R. China}%
\thanks{$^{2}$Zhejiang University,
        Zhejiang, 310058, P.R. China}%
\thanks{$^{\dag}$Corresponding author is also affiliated with School of Artificial Intelligence, 
        {\tt\small ruihuang@cuhk.edu.cn}}%
}

\begin{document}

\maketitle
\thispagestyle{empty}
\pagestyle{empty}

\begin{abstract}
We present a novel approach, termed ADGaussian, for generalizable street scene reconstruction. The proposed method enables high-quality rendering from merely single-view input. 
Unlike prior Gaussian Splatting methods that primarily focus on geometry refinement, we emphasize the importance of joint optimization of image and depth features for accurate Gaussian prediction. To this end, we first incorporate sparse LiDAR depth as an additional input modality, formulating the Gaussian prediction process as a joint learning framework of visual information and geometric clue. Furthermore, we propose a Multi-modal Feature Matching strategy coupled with a Multi-scale Gaussian Decoding model to enhance the joint refinement of multi-modal features, thereby enabling efficient multi-modal Gaussian learning. Extensive experiments on Waymo and KITTI demonstrate that our ADGaussian achieves state-of-the-art performance and exhibits superior zero-shot generalization capabilities in novel-view shifting. \href{https://maggiesong7.github.io/research/ADGaussian/}{Project page.}

\end{abstract}

\section{INTRODUCTION}

Recently, 3D Gaussian Splatting (3DGS) \cite{kerbl20233d} has garnered significant attention in the fields of 3D scene reconstruction and novel view synthesis \cite{mildenhall2021nerf} due to its real-time rendering speed and high-quality output.
One key application is the modeling of street scenes from image sequences, which plays a vital role in areas such as autonomous driving \cite{song2024divide}.

\begin{figure*}[!t]
\centering
\includegraphics[width=0.94\textwidth]{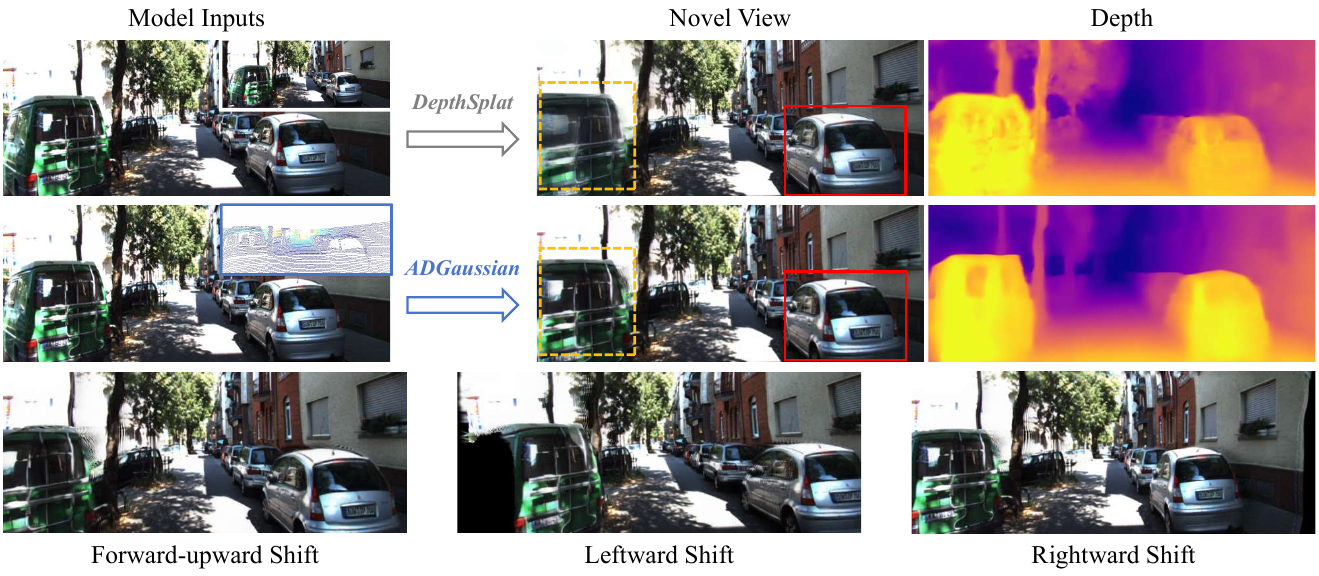} 
\vspace{-3mm}
\caption{We introduce \textbf{ADGaussian}, a generalizable Gaussian framework that achieves superior street scene reconstruction in both visual and geometric quality. The bottom row illustrates the results of viewpoint shifting, further demonstrating the robustness of our method under varying viewpoint changes.}
\vspace{-2mm}
\label{fig:teaser}
\end{figure*}

When modeling urban scenes, some methods follow per-scene optimization techniques \cite{kung2024lihi, yan2025streetcrafter, zhao2025recondreamer++}, notably StreetGaussians \cite{yan2023street} that represents dynamic urban street as a set of point clouds equipped with semantic logits and 3D Gaussians. While such an approach excels in high-quality reconstruction, it struggles with expensive training cost and large-range novel view synthesis, which motivates the exploration of generalizable models that avoid scene-specific fine-tuning.

To achieve generalizable street scene reconstruction, most existing methods build upon the architectures of PixelSplat \cite{charatan2024pixelsplat} or MVSplat \cite{chen2024mvsplat}. For instance, GGRt \cite{li2024ggrt} introduces a pose-free architecture to iteratively update multi-view depth map and subsequently estimates Gaussian primitives based on PixelSplat. Similarly, GGS \cite{han2024ggs} enhances the depth estimations of MVSplat by integrating a multi-view depth refinement module. Nevertheless, multi-view feature matching-based depth estimation may fail in challenging conditions such as texture-less areas and reflective surfaces. 
To tackle this issue, the concurrent work DepthSplat \cite{xu2024depthsplat} combines pre-trained depth features from Depth Anything V2 \cite{yang2024depth} with multi-view depth estimations for accurate depth regression, where the estimated depth features are then concatenated with image features for Gaussian prediction.

Given the great generalization capability of Depth Anything V2, it is reasonable to extend DepthSplat to urban street scenarios. However, DepthSplat faces specific limitations when applied to these environments.  
First, as highlighted in the dashed box of Fig. \ref{fig:teaser}, the visual rendering quality is constrained by the performance of pre-trained depth models, which often exhibit inconsistent accuracy across diverse street datasets and scenarios.    
Second, simply concatenating image and depth features for the final Gaussian prediction without any information sharing or multi-modal feature fusion may lead to unexpected spatial misalignment, evidenced by the distorted car shape in the red box of Fig. \ref{fig:teaser}.

To overcome these limitations, we present ADGaussian, a novel multi-modal framework that jointly optimizes visual rendering quality and geometric accuracy for street scenes. Instead of relying on pre-trained depth foundation models, we choose to integrate sparse LiDAR depth information as an additional input modality, which is more practical in real-world scenarios and provides precise metric scale priors for geometry reconstruction. Given two complementary modalities (i.e., image data and sparse depth map), the framework's core innovation lies in its synergistic processing of these two data streams, enabling effective information sharing and joint optimization between different modalities. Specifically, we introduce a Multi-modal Feature Matching strategy augmented by the Depth-guided Position Embedding, which contains a Siamese-style encoder paired with an information cross-attention decoder. This design ensures a cohesive fusion of geometric and appearance information, resulting in well-aligned multi-modal tokens. 
Subsequently, we employ a Multi-scale Gaussian Decoding model to aggregate multi-scale depth information into the resulting multi-modal tokens for the final 3D Gaussian predictions.

As presented in Fig. \ref{fig:teaser}, ADGaussian excels in both visual rendering and geometric reconstruction. Notably, the bottom row demonstrates our model's robustness under extreme viewpoint variations, where competing methods often produce distorted geometry. This superior performance validates the efficacy of ADGaussian's synchronized cross-modal optimization paradigm.

Overall, this work makes the following contributions:
\begin{itemize}
\setlength{\itemsep}{0pt}
\setlength{\parsep}{0pt}
\setlength{\parskip}{0pt}
\item We present ADGaussian, the first generalizable framework that formulates street scene Gaussian prediction as joint visual and geometric learning. 
\item We develop a Multi-modal Feature Matching strategy along with a Multi-scale Gaussian Decoding model to facilitate effective multi-modal Gaussian learning.
\item We conduct extensive comparisons on two datasets, verifying our approach’s state-of-the-art performance and the effectiveness of the proposed components.
\end{itemize}

\section{RELATED WORK}

\subsection{Generalizable 3D Gaussian Splatting}

Generalizable Gaussian Splatting \cite{wang2024freesplat, wewer2024latentsplat, chen2024g3r} aims to learn powerful priors that enable effective generalization across unseen scenes. Existing methods can be broadly categorized into two groups based on their handling of camera parameters. The first group, including approaches like MVSGaussian \cite{liu2024mvsgaussian} and SplatterImage \cite{szymanowicz2024splatter}, predicts per-pixel 3D Gaussian primitives using known camera parameters. These works typically demonstrate superior reconstruction accuracy due to the precise geometric constraints provided by the camera poses.
The second group of methods \cite{smart2024splatt3r, ye2024no} proposes to jointly predict camera parameters and 3D representations, eliminating the need for known camera poses. For instance, GGRt \cite{li2024ggrt} employs an Iterative Pose Optimization Network to estimate and iteratively update the relative pose between target and reference images. DrivingForward \cite{tian2025drivingforward} adopts pose network and depth network to determine the position of the Gaussian primitives in a self-supervised manner.
In street scene modeling, however, camera poses provide critical constraints for determining scene scale and enhancing reconstruction accuracy from image sequences. Moreover, camera poses are readily accessible in street scenes, making them a practical and reliable data resource. Therefore, we choose to leverage posed images for our approach. 

\subsection{Depth and Gaussian Splatting}

Depth quality has been demonstrated to play a pivotal role in Gaussian Splatting, serving as the foundation for accurate geometry reconstruction and realistic rendering. To ensure precise geometric fidelity, existing approaches \cite{lu2024drivingrecon, zheng2024gps, turkulainen2024dn} incorporate additional depth supervision into the optimization process. However, since dense Ground Truth depth data is often unavailable in practical applications, researchers have turned to pre-trained depth foundation models \cite{piccinelli2024unidepth, yang2024depth} as an alternative source of reliable geometric cues. For example, Chung et al. \cite{chung2024depth} rescale pre-trained depth maps using sparse COLMAP points to provide depth constraints. DepthSplat \cite{xu2024depthsplat} fuses pre-trained depth features with multi-view cost volume features to help depth refinement. While these methods have significantly improved geometric accuracy, we contend that their primary focus on geometry enhancement overlooks the crucial interplay between appearance and structure. In contrast to previous approaches, we argue that joint optimization of image and depth features is more critical for achieving high-quality reconstruction that excels in both visual fidelity and structural accuracy.

\subsection{LiDAR-Integrated Gaussian Splatting}

The integration of LiDAR data has emerged as a widely adopted approach in street scene reconstruction \cite{wang2023cadsim, yang2023reconstructing}, due to its effectiveness in facilitating geometry learning. 
The conventional methodologies typically involve two main steps: initializing Gaussians from LiDAR point clouds \cite{khan2024autosplat} to establish the basic scene structure, and further supervising predicted Gaussian positions with LiDAR priors \cite{jiang2024li, huang2024textit} to refine geometric details.
Typically, TCLC-GS \cite{zhao2024tclc} constructs a hybrid 3D representation by combining LiDAR geometries with image colors, enabling simultaneous initialization of both geometric and appearance attributes of 3D Gaussians. 
Rather than directly using LiDAR point clouds, we propose leveraging sparse LiDAR depth to bridge the modality gap between LiDAR and camera data. Furthermore, we integrate depth priors as an additional input modality into a unified optimization framework, achieving joint optimization of depth geometries and image photometric attributes, as opposed to the common practice of initialization alone.

\begin{figure*}[ht]
\centering
   \includegraphics[width=1.\textwidth]{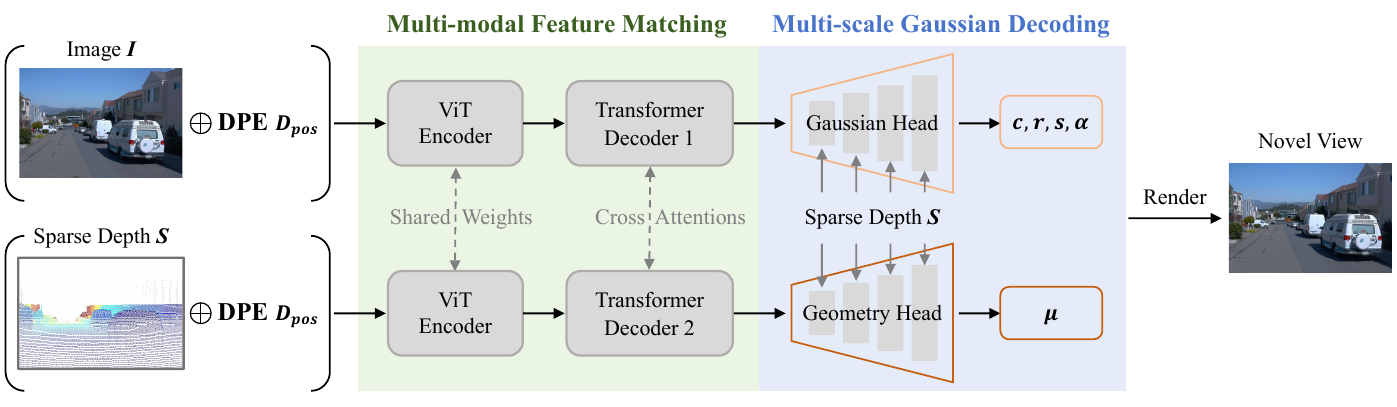}
   \vspace{-6.2mm}
   \caption{\textbf{Overall framework of ADGaussian.} Given monocular posed image and sparse depth as inputs, we first extract fused multi-modal features through Multi-modal Feature Matching strategy enhanced by Depth-guided Positional Embedding (DPE). These aligned multi-modal tokens are then processed by our Multi-scale Gaussian Decoding module, which hierarchically integrates depth cues across scales to produce optimized 3D Gaussian outputs.}
   \label{overall_arc}
\end{figure*}

\vspace{-1.3mm}
\section{Method}

Depth foundation models \cite{piccinelli2024unidepth, yang2024depth} have been integrated into the Gaussian Splatting to improve geometry reconstruction. However, such a framework often suffers from suboptimal rendering quality due to the insufficient interactions between photometric and geometric clues. To address this, we propose ADGaussian, a synchronized multi-modal optimization architecture that combines sparse depth data with monocular images for enhanced street scene modeling. 

\vspace{-1.2mm}
\subsection{Preliminary}

Recently, some works have investigated the advantages of using a pre-trained depth foundation model for image-conditioned 3D Gaussian reconstruction. 
For instance, DepthSplat \cite{xu2024depthsplat} processes multi-view images \(\{I^i\}^N_{i=1}(I\in \mathbb{R}^{H \times W \times 3})\) using two parallel branches to extract dense per-pixel depth. One branch focuses on modeling cost volume features \(C^i\) from the multi-view input, while the other employs a pre-trained monocular depth backbone, specifically Depth Anything V2 \cite{yang2024depth}, to obtain monocular depth features \(F^i_{mono}\). These per-view cost volumes and monocular features are then concatenated for depth regression. Finally, DepthSplat predicts all remaining Gaussian parameters using the concatenated image, depth, and feature information.

Intuitively, such models can be easily adapted to urban scenes. Nonetheless, we observed that the effectiveness of reconstruction is heavily dependent on the performance of the pre-trained depth foundation models, resulting in inconsistent accuracy across different street datasets and scenarios. Furthermore, the processing of image and depth features always occurs in parallel for each view, without any information sharing or synchronized optimization, which constrains the model's learning capacity. 

In this paper, instead of relying on pre-trained depth foundation models, we propose utilizing sparse LiDAR depth measurements as an additional input. This choice is particularly advantageous for street scene applications, as LiDAR data offers both greater practical accessibility in real-world autonomous driving systems and reliable metric-scale depth priors crucial for accurate geometric reconstruction.

\vspace{-1.1mm}
\subsection{Multi-modal Feature Matching}
In this subsection, we seek to find an effective way to integrate sparse LiDAR depth into Gaussian Splatting. To this end, we propose a \textit{Multi-modal Feature Matching} architecture tailored for urban scenarios to enable the synchronous integration of sparse depth information and color image data. Throughout this process, \textit{Depth-guided Position Embedding} incorporates depth cues into the position embedding, enhancing 3D spatial awareness and improving multi-modal contextual comprehension.

\begin{table*}[htp]
\caption{\textbf{Quantitative comparisons with state of the art on Waymo dataset}. Our ADGaussian outperforms existing methods in nearly all scenarios. Cells highlighted in {\begin{tikzpicture} [scale=1.0] \fill[deeppeach] (0,0) rectangle (1em,0.8em); \end{tikzpicture}} {\begin{tikzpicture}[scale=1.0]  \fill[color=lightgray] (0ex,0ex) rectangle (1em,0.8em);  \end{tikzpicture} denotes the best and second-best performances.} }
\vspace{-1mm}
\centering
\resizebox{1.0\textwidth}{!}{
\begin{tabular}{clccccccccc}
\toprule
\multicolumn{2}{c}{\multirow{2}{*}{Scene}}       & \multicolumn{3}{c}{PSNR\(\uparrow\)}                        & \multicolumn{3}{c}{SSIM\(\uparrow\)}                        & \multicolumn{3}{c}{LPIPS\(\downarrow\)}                        \\ \cmidrule(r){3-5}\cmidrule(r){6-8}\cmidrule(r){9-11}
\multicolumn{2}{c}{}                             & MVSplat \cite{chen2024mvsplat} & \multicolumn{1}{c}{DepthSplat \cite{xu2024depthsplat}} & \textbf{Ours} & MVSplat \cite{chen2024mvsplat} & \multicolumn{1}{c}{DepthSplat \cite{xu2024depthsplat}} & \textbf{Ours} & MVSplat \cite{chen2024mvsplat} & \multicolumn{1}{c}{DepthSplat \cite{xu2024depthsplat}} & \textbf{Ours} \\ \midrule
\multirow{4}{*}{\rotatebox{90}{Static}}                     & 003   &         \cellcolor{lightgray}21.79& \multicolumn{1}{c}{19.99}           &      \cellcolor{deeppeach}31.09&         \cellcolor{lightgray}0.679& \multicolumn{1}{c}{0.627}           &      \cellcolor{deeppeach}0.931&         \cellcolor{lightgray}0.143& \multicolumn{1}{c}{0.192}           &      \cellcolor{deeppeach}0.059\\
                                           & 069   &         24.79& \multicolumn{1}{c}{\cellcolor{lightgray}25.67}           &      \cellcolor{deeppeach}31.17&         0.729& \multicolumn{1}{c}{\cellcolor{lightgray}0.748}           &      \cellcolor{deeppeach}0.923&         0.143& \multicolumn{1}{c}{\cellcolor{lightgray}0.136}           &      \cellcolor{deeppeach}0.073\\
                                           & 232  &         \cellcolor{lightgray}28.79& \multicolumn{1}{c}{26.76}           &      \cellcolor{deeppeach}30.52&         \cellcolor{lightgray}0.873& \multicolumn{1}{c}{0.819}           &      \cellcolor{deeppeach}0.904&         \cellcolor{deeppeach}0.077& \multicolumn{1}{c}{0.094}           &      \cellcolor{lightgray}0.083\\
                                           & 495  &         \cellcolor{lightgray}28.09& \multicolumn{1}{c}{26.49}           &      \cellcolor{deeppeach}31.21&         \cellcolor{lightgray}0.884& \multicolumn{1}{c}{0.819}           &      \cellcolor{deeppeach}0.929&         \cellcolor{lightgray}0.086& \multicolumn{1}{c}{0.106}           &      \cellcolor{deeppeach}0.056\\ \midrule
\multicolumn{1}{l}{\multirow{5}{*}{\rotatebox{90}{Dynamic}}} & 016 &   24.16      & \multicolumn{1}{c}{\cellcolor{lightgray}24.30}           &      \cellcolor{deeppeach}27.16&    0.678     & \multicolumn{1}{c}{\cellcolor{lightgray}0.746}           &      \cellcolor{deeppeach}0.875&    \cellcolor{lightgray}0.137     & \multicolumn{1}{c}{0.173}           &      \cellcolor{deeppeach}0.092\\
\multicolumn{1}{l}{}                       & 021 &    \cellcolor{lightgray}19.58     & \multicolumn{1}{c}{18.42}           &      \cellcolor{deeppeach}19.61&    \cellcolor{lightgray}0.636     & \multicolumn{1}{c}{0.619}           &      \cellcolor{deeppeach}0.659&    \cellcolor{deeppeach}0.243     & \multicolumn{1}{c}{0.316}           &      \cellcolor{lightgray}0.273\\
\multicolumn{1}{l}{}                       & 080 &    \cellcolor{lightgray}25.37     & \multicolumn{1}{c}{24.19}           &\cellcolor{deeppeach}27.18&    \cellcolor{lightgray}0.765     & \multicolumn{1}{c}{0.759}           &      \cellcolor{deeppeach}0.873&     \cellcolor{lightgray}0.116    & \multicolumn{1}{c}{0.169}           &      \cellcolor{deeppeach}0.085\\
\multicolumn{1}{l}{}                       & 096 &    \cellcolor{lightgray}21.55    & \multicolumn{1}{c}{\cellcolor{deeppeach}21.67}           &      21.46&   \cellcolor{lightgray}0.684      & \multicolumn{1}{c}{0.680}           &     \cellcolor{deeppeach}0.691&   \cellcolor{deeppeach}0.250      & \multicolumn{1}{c}{0.264}           &      \cellcolor{lightgray}0.263\\\bottomrule
\end{tabular}}
\vspace{-2mm}
\label{sota_w}
\end{table*}

\paragraph{Multi-modal Feature Matching} 

As illustrated in Fig. \ref{overall_arc}, the first part of our model is the \textit{Multi-modal Feature Matching} of photometric features from the image and geometric cues from depth data. This is achieved through a Siamese-style encoder and an information cross-attention decoder, inspired by the DUSt3R series \cite{wang2024dust3r, leroy2024grounding}.


Specifically, given a monocular image \(I\in \mathbb{R}^{H \times W \times 3}\) and synchronized sparse depth map \(S\in \mathbb{R}^{H \times W \times 1}\), we first replicate the depth map across channels to match the image's dimensional structure. These multi-modal inputs are then fed into a weight-sharing ViT encoder, resulting in two token representations \(F_I\) and \(F_S\):
\begin{equation}
  F_I=\text{Encoder}(I), F_S=\text{Encoder}(S)
  \label{lossfunc1}
\end{equation}
The two identical encoders collaboratively process multi-modal features in a weight-sharing manner, allowing for the automatic learning of similarity characteristics.

After that, the transformer decoders equipped with cross attentions are employed to enhance information sharing and synchronized optimization between the two multi-modal branches. This step is crucial for producing well-fused multi-modal feature maps:
\begin{equation}
  \begin{aligned}
      G_I&=\text{Decoder}_1(F_I,F_S), \\
      G_S&=\text{Decoder}_2(F_S,F_I)
  \end{aligned}
\end{equation}

\paragraph{Depth-guided Positional Embedding (DPE)} 
The conventional positional embedding in Vision Transformers encodes either relative or absolute spatial positions on a 2D image plane to ensure spatial awareness within the image. However, relying solely on the geometric properties of a 2D image plane is insufficient for our synchronized multi-modal design. To this end, we propose a straightforward Depth-guided Positional Embedding (DPE) to integrate depth positions with image-based spatial positions

In particular, given the downsampled image size \(H_L\times W_L\) and the sparse depth map, we first flatten the 2D grid of spatial positions into a 1D vector through row-major ordering, where each spatial position \((i,j)\) in the 2D grid is mapped to linear index \(k=i\times W_L+j\) in the 1D vector. Meanwhile, the sparse depth map is downsampled to match the image resolution, generating an independent set of depth indices that complement the spatial positions. The final positional embedding \(D_{pos}\) is then formed by concatenating the flattened spatial coordinates with their corresponding depth values, establishing an integrated \textit{xy}-\textit{z} coordinate representation that encodes both planar and depth-wise positional information. 
By integrating both spatial and depth geometry, this module provides a comprehensive positional prior for effective multi-modal feature fusion.

\subsection{Multi-scale Gaussian Decoding}

Given the multi-modal tokens \(G_I\) and \(G_S\), our objective is to predict pixel-aligned Gaussian parameters \(\{(\mu,\alpha,\Sigma,c)\}^{H\times W} \), where \(\mu\), \(\alpha\), \(\Sigma\), and \(c\) are the 3D Gaussian’s center position, opacity, covariance, and color information. To fully leverage appearance cues and the geometry priors provided by image token \(G_I\) and depth token \(G_S\), we implement two separate regression heads with the same architecture, namely Gaussian Head and Geometry Head, to generate different Gaussian parameters. 

The two regression heads adhere to the DPT \cite{ranftl2021vision} architecture, enhanced with an additional multi-scale depth encoding that delivers precise scale priors for Gaussian prediction. In particular, at each scale within the DPT Decoder, we initially resize the input depth map to align with the spatial size of the current feature scale. After that, the resized depth map is processed through a shallow network comprising two convolutional layers to extract depth features, which are then added to the DPT intermediate features. 
Finally, the input image and depth map, each processed by a single convolutional layer, are individually incorporated into the final features of the Gaussian Head and Geometry Head to facilitate either appearance-based or geometry-based Gaussian decoding.

\subsection{Training Loss}
Our model is trained using a combination of novel view synthesis loss and depth loss:
\begin{equation}
  \mathcal{L}=\mathcal{L}_{\text{nvs}} + \mathcal{L}_{\text{depth}} 
  \label{lossfunc1}
\end{equation}

\paragraph{Novel view synthesis loss} We train our full model with a combination of mean squared error (MSE) and LPIPS losses between rendered and Ground Truth image colors:
\begin{equation}
  \mathcal{L}_{\text{nvs}} = \text{MSE}(I_{\text{pred}}, I_{\text{gt}}) + \lambda \cdot\text{LPIPS}(I_{\text{pred}}, I_{\text{gt}})
  \label{lossfunc2}
\end{equation}
where the LPIPS loss weight \(\lambda\) is set to 0.05.

\paragraph{Depth loss} We leverage depth loss to smooth the depth values of neighboring pixels, thereby minimizing abrupt changes over small regions:
\begin{equation}
  \mathcal{L}_{\text{depth}} = \frac{1}{n}\sum_{i=1}^n(\frac{dD_i}{dx}e^{-\frac{dI_i}{dx}}+\frac{dD_i}{dy}e^{-\frac{dI_i}{dy}})
  \label{lossfunc3}
\end{equation}
where \(\frac{dD_i}{dx}\), \(\frac{dD_i}{dy}\), \(\frac{dI_i}{dx}\), and \(\frac{dI_i}{dy}\) denote the first derivatives of depth and image in the x and y-axis directions, respectively.

\section{Experiments}

\subsection{Implementation Details}

\paragraph{Datasets and metrics}  
We evaluate our proposed approach on two widely used autonomous driving datasets: the Waymo Open Dataset \cite{sun2020scalability} and the KITTI Tracking benchmarks \cite{geiger2012we}. For both datasets, we adopt a train-test split ratio of approximately 1:7. Specifically, on the Waymo dataset, our focus primarily lies on static and dynamic scenes, where each scene type is divided into 4 test scenes and 28 training scenes. Similarly, for the KITTI dataset, the split consists of 5 test scenes and 37 training scenes. For render quality evaluation, we employ the standard image quality metrics, including Peak Signal-to-Noise Ratio (PSNR),
Structural Similarity Index Measure (SSIM) \cite{wang2004image}, and the Learned Perceptual Image Patch Similarity (LPIPS) \cite{zhang2018unreasonable}. 


\begin{figure*}[ht]
\centering
   \includegraphics[width=1.0\textwidth]{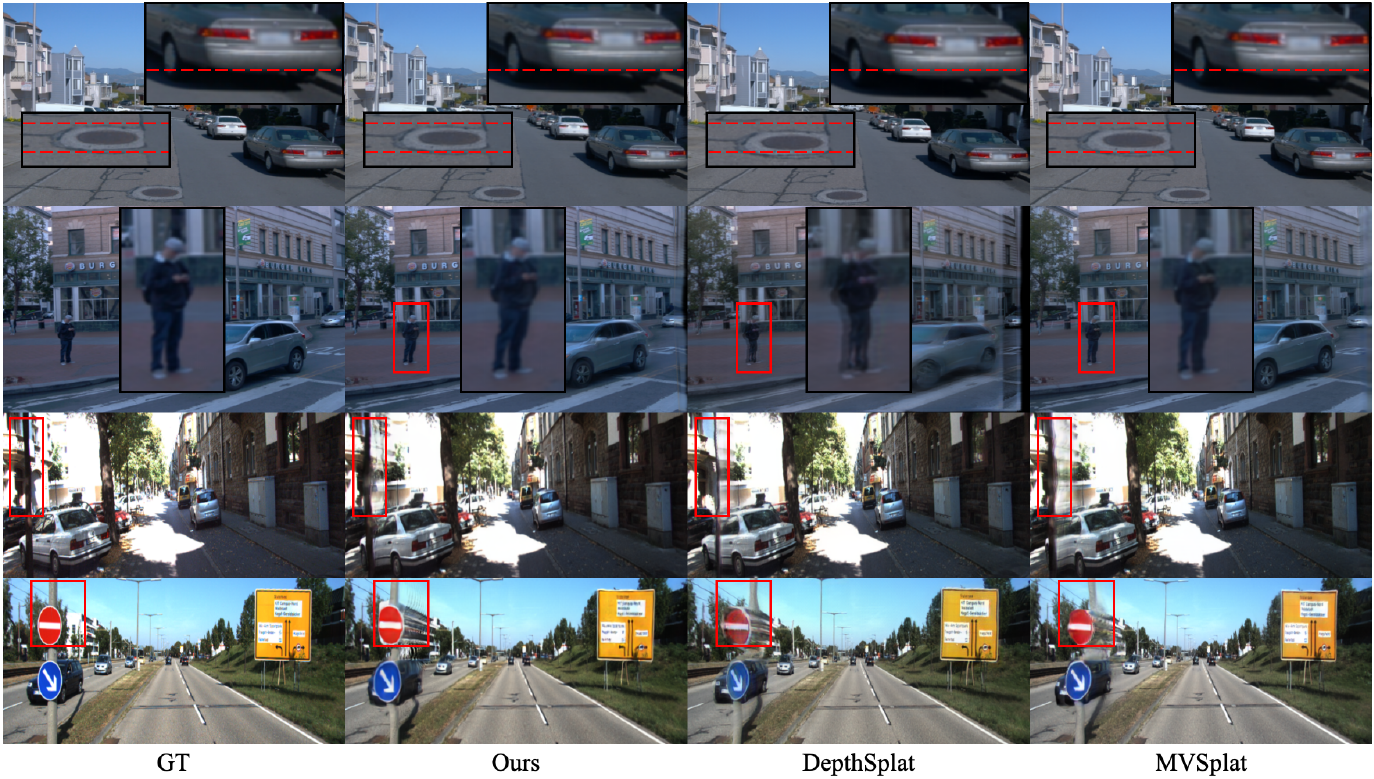}
   \vspace{-6mm}
   \caption{\textbf{Qualitative comparisons with state of the art.} Our ADGaussian surpasses all other competitive models in rendering quality within urban scenarios. The zoom-in comparisons in the first row further reveal our enhanced spatial alignment capabilities. }
   \vspace{-2mm}
   \label{vis_sotaw}
\end{figure*}


\paragraph{Training details} We employ the Adam optimizer and cosine learning rate schedule, with an initial learning rate of \(1e\)-\(4\). We train our model on a single 3090 Ti GPU, running for 150k iterations on both Waymo and KITTI datasets, with a batch size of 1. To ensure a fair comparison, all experiments are carried out at resolutions of 320\textit{×}480 for the Waymo dataset and 256\textit{×}608 for the KITTI dataset.

\subsection{Comparisons with the State of the Art}

When comparing our work with current state-of-the-art Gaussian Splatting methods, we selected MVSplat \cite{chen2024mvsplat} (a multi-view cost volume-based approach) and DepthSplat \cite{xu2024depthsplat} (a depth foundation model-based approach) as primary baselines. Since our method focuses on pose-aware generalizable Gaussian Splatting, we excluded pose-free methods and per-scene optimized baselines from the main comparisons.
To ensure fair and consistent comparisons, all baseline methods were re-implemented and trained on both datasets using identical experimental settings to ours. Specifically, for each scenario, both MVSplat and DepthSplat utilize consecutive frame pairs as input, with the subsequent immediate frame serving as the target novel view for evaluation.

 

\begin{table}[t]
\caption{\textbf{Quantitative comparisons with state of the art on KITTI dataset.} KITTI¹ refers to subsequent temporal frame rendering, while KITTI² indicates spatial left-to-right camera rendering. Cells highlighted in {\begin{tikzpicture} [scale=1.0] \fill[deeppeach] (0,0) rectangle (1em,0.8em); \end{tikzpicture}} {\begin{tikzpicture}[scale=1.0]  \fill[color=lightgray] (0ex,0ex) rectangle (1em,0.8em);  \end{tikzpicture} denotes the best and second-best performances.}}
\vspace{-1mm}
\centering
\resizebox{1.01\columnwidth}{!}{
\begin{tabular}{lcccccc}
\toprule
\multirow{2}{*}{Method}& \multicolumn{3}{c}{KITTI\(^1\)}     & \multicolumn{3}{c}{KITTI\(^2\)} \\ \cmidrule(r){2-4}\cmidrule(r){5-7}
                        & PSNR\(\uparrow\)  & SSIM\(\uparrow\) & LPIPS\(\downarrow\) & PSNR\(\uparrow\)   & SSIM\(\uparrow\)  & LPIPS\(\downarrow\)  \\ \midrule
MVSplat \cite{chen2024mvsplat} & \cellcolor{lightgray}23.52      &    \cellcolor{lightgray}0.760        &  \cellcolor{deeppeach}0.152   &24.65&0.833& 0.141 \\
DepthSplat \cite{xu2024depthsplat} &   21.99     &     0.715       &   0.173   &\cellcolor{lightgray}25.36&\cellcolor{lightgray}0.838& \cellcolor{lightgray}0.135 \\
Ours &  \cellcolor{deeppeach}23.60     &    \cellcolor{deeppeach}0.776      &   \cellcolor{lightgray}0.164  &\cellcolor{deeppeach}26.10&\cellcolor{deeppeach}0.853& \cellcolor{deeppeach}0.122 \\ \bottomrule
\end{tabular}}
\vspace{-2mm}
\label{sota_k}
\end{table}

The quantitative comparisons on the Waymo and KITTI benchmarks are presented in Table \ref{sota_w} and Table \ref{sota_k}, respectively. On the Waymo dataset, our ADGaussian surpasses previous state-of-the-art models on almost all visual metrics, with particularly significant gains in static scenes and consistent performance across diverse scenarios. 
While competing methods suffer from pervasive spatial misalignment issues that severely degrade their metric accuracy (as evidenced in Fig. \ref{vis_sotaw}), our model effectively resolves this fundamental challenge through cross-modal joint learning, which synchronizes geometric and appearance optimization to achieve pixel-perfect alignment. On the KITTI dataset, we introduce an additional left-to-right view synthesis setting, leveraging its stereo camera data. As shown in Table \ref{sota_k}, our method achieves superior performance in both settings compared to prior works. However, the performance gain on KITTI is less pronounced compared to that on Waymo. This is primarily attributed to the overall lower image quality and poor color reproduction of the KITTI dataset. Since our method relies solely on a single image as input, it retains fewer image details compared to previous works, which further constrains its performance on datasets with inferior image quality.

We provide qualitative results of the two datasets in Fig. \ref{vis_sotaw}. As can be observed, the zoomed regions in the first row reveal pervasive spatial misalignment in previous methods, which leads to significant degradation in evaluation metrics. The last two rows further demonstrate that our model achieves superior rendering quality, particularly in occluded regions and fine details such as slender signal poles.

Besides, the comparison between DepthSplat and MVSplat shows that DepthSplat exhibits stronger depth inference capabilities, attributed to its enhanced geometry reconstruction facilitated by pre-trained depth models. However, DepthSplat falls short in overall visual reconstruction quality due to its insufficient integration of appearance attributes, which is consistent with our earlier analysis in the preceding sections.

\subsection{Ablations and Analyses}

\paragraph{Ablations on proposed components}

\begin{table}[t]
\caption{\textbf{Ablation studies on the Waymo dataset.} We report the averaged scores across all validation scenes for a more intuitive reflection of model performance.}
\vspace{-1mm}
\centering
\resizebox{0.9\columnwidth}{!}{
\begin{tabular}{lccc}
\toprule
Setup & PSNR\(\uparrow\) & SSIM\(\uparrow\) & LPIPS\(\downarrow\) \\
\midrule
Full Model & 31.00& 0.921&0.068\\ \midrule
w/o DPE & 30.31& 0.908& 0.078\\
w/o Multi-scale& 28.73& 0.868&0.100\\
w/o DPE \& Multi-scale & 27.81& 0.846& 0.114\\
w/o Matching & 26.68& 0.814& 0.106\\
\bottomrule
\end{tabular}}
\label{ablation1}
\end{table}

\begin{table}[t]
\caption{\textbf{Analyses on forward view shifting and running efficiency.}}
\vspace{-1mm}
\centering
\resizebox{0.95\columnwidth}{!}{
\begin{tabular}{lccccc}
\toprule
Method & PSNR\(\uparrow\) & SSIM\(\uparrow\) & LPIPS\(\downarrow\) &Time\(\downarrow\)&Memory\(\downarrow\)\\
\midrule
MVSplat \cite{chen2024mvsplat} &  24.84   &   0.777   & 0.133 &0.22/0.14& 11.11 \\
DepthSplat \cite{xu2024depthsplat} &  23.40   &   0.726   & 0.190 & 0.37/0.28& 21.17 \\
Ours  &  27.68   &  0.877    & 0.101 & 0.29/0.18&17.52 \\
\bottomrule
\end{tabular}}
\vspace{-2mm}
\label{ablation_forward}
\end{table}

\begin{figure}[tp]
\centering
   \includegraphics[width=1.0\columnwidth]{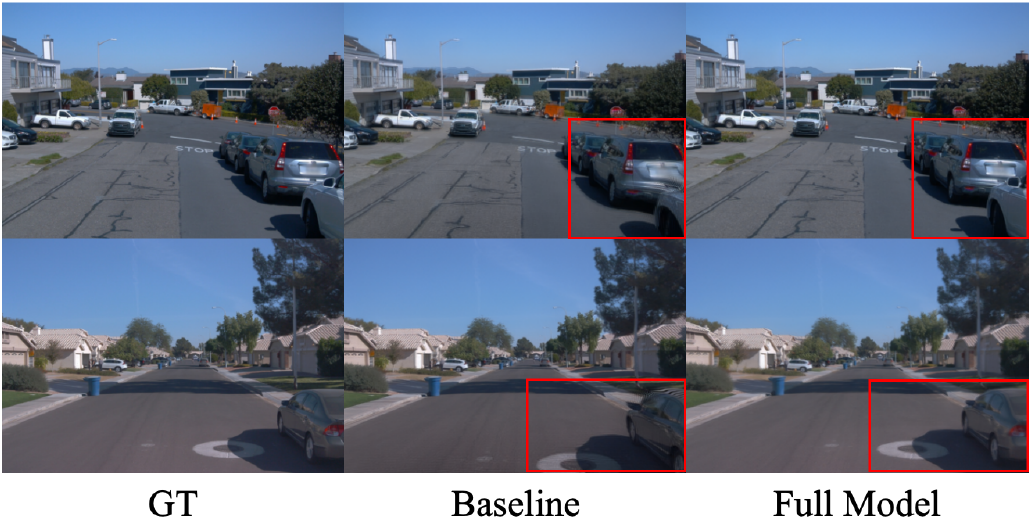}
   \vspace{-6mm}
   \caption{\textbf{Ablations on the Waymo dataset.} We present a visual comparisons between our baseline model (w/o DPE \& Multi-scale) and the full model. Our experiments demonstrate that the full model can effectively address the spatial misalignment issues prevalent in street scene reconstruction.}
   \vspace{-2mm}
   \label{vis_ablation}
\end{figure}

\begin{figure}[t]
\centering
   \includegraphics[width=1.0\columnwidth]{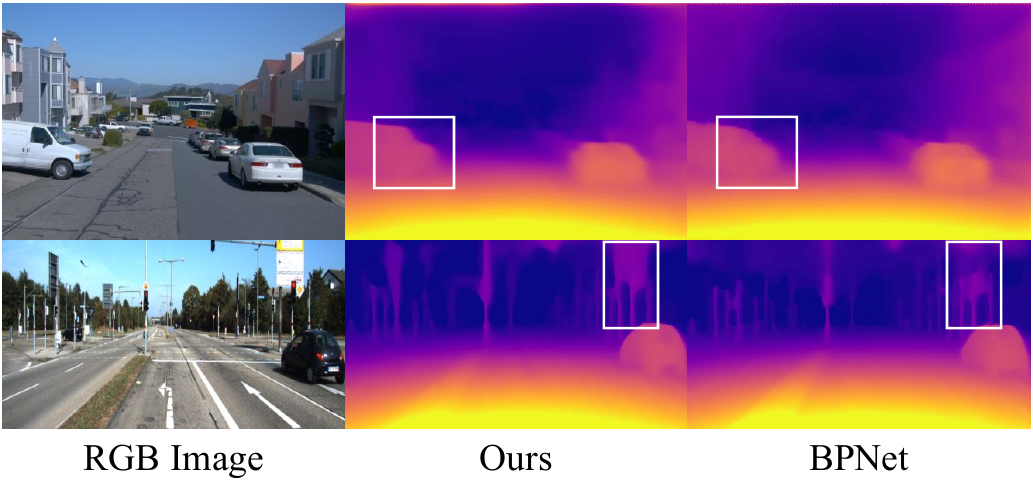}
   \vspace{-6mm}
   \caption{\textbf{Depth comparisons with depth completion networks.} Our method demonstrates superior depth estimation performance in certain challenging regions, even without depth pre-training.}
   \vspace{-2mm}
   \label{vis_depth}
\end{figure}


The ablation studies are detailed in Table \ref{ablation1} to confirm the efficacy of proposed components.
First, it can be seen that the full model achieves the highest performance, boasting a PSNR, SSIM, and LPIPS score of 31.0, 0.921, and 0.068, respectively. Notably, the removal of the Depth-guided Positional Embedding (DPE) resulted in a decrease across all metrics (0.69, 1.3\%, and 1\%, respectively), emphasizing the significance of depth positions in facilitating the joint optimization of multi-modal features. Furthermore, the model lacking Multi-scale Gaussian Decoding (w/o Multi-scale) exhibited reduced performance, achieving a PSNR of 28.73, an SSIM of 0.868, and an LPIPS of 0.100, underscoring the effectiveness of multi-level depth decoding and independent Gaussian inference. Removing both DPE and Multi-scale led to a more substantial drop in performance, notably a 4.6\% decrease in the LPIPS score. This quantitative degradation aligns with the qualitative results in Fig. \ref{vis_ablation}, which validates the importance of our DPE and Multi-scale Gaussian Decoding in addressing spatial misalignment for street scene reconstruction.

Finally, to showcase the effectiveness of our synchronized multi-modal optimization formulation, we present the results without Multi-modal Feature Matching (w/o Matching) by substituting the sparse depth input with a color image from the subsequent frame. It is evident that Multi-modal Feature Matching brought about significant enhancement in PSNR, SSIM, and LPIPS (4.32, 10.7\%, and 3.8\%, respectively), highlighting the importance of information exchange and synchronized optimization of image-related appearance features and depth-related geometric features.


\paragraph{Robustness across frames}  
In the previous experiments, the next frame (\textit{t+1} frame) was used as the target for novel view synthesis. To further evaluate the robustness under more significant viewpoint changes, we extend the prediction to the \textit{t+2} frame. As reported in Table \ref{ablation_forward}, our ADGaussian outperforms in handling larger temporal and spatial shifts, even with only a single input frame.

\paragraph{Analysis on running efficiency}
The right side of table \ref {ablation_forward} shows training/inference runtime and memory consumption of ADGaussian, MVSplat and DepthSplat at 320×480 resolution to validate ADGaussian’s practicality. ADGaussian uses 3.65 GB less training memory than DepthSplat for higher efficiency, and its inference speed is merely 0.04 s slower than MVSplat with comparable overall runtime.

\begin{table*}[t]
\caption{\textbf{Performance analysis on multi-modal inputs.} Models marked with "\(*\)" are modified with a Gaussian head for 3DGS prediction. The term "Waymo + Depth Drop" refers to our robustness evaluation on depth quality.}
\vspace{-1mm}
\centering
\resizebox{0.85\textwidth}{!}{
\begin{tabular}{lccccccccc}
\toprule
\multirow{2}{*}{Method}& \multicolumn{3}{c}{Waymo}     & \multicolumn{3}{c}{KITTI} & \multicolumn{3}{c}{Waymo + Depth Drop}\\ \cmidrule(r){2-4}\cmidrule(r){5-7}\cmidrule(r){8-10}
                        & PSNR\(\uparrow\)  & SSIM\(\uparrow\) & LPIPS\(\downarrow\) & PSNR\(\uparrow\)   & SSIM\(\uparrow\)  & LPIPS\(\downarrow\) & PSNR\(\uparrow\)   & SSIM\(\uparrow\)  & LPIPS\(\downarrow\)  \\ \midrule
CFormer\(^*\) \cite{zhang2023completionformer}                    &   25.71  & 0.796  &  0.126    &      21.35   &  0.761 &  0.192   &23.67&0.780&0.141\\
   BPNet\(^*\) \cite{tang2024bilateral}                 &    26.10 &0.802  & 0.144     &   19.68 & 0.626& 0.336 &24.89&0.783& 0.152\\
  Ours                 &    31.00& 0.921&0.068   &   23.60 &0.776 &0.164 &30.56&0.912&0.074\\ \bottomrule
\end{tabular}}
\vspace{-1mm}
\label{ablation2}
\end{table*}


\begin{figure*}[htp]
\centering
   \includegraphics[width=1.0\textwidth]{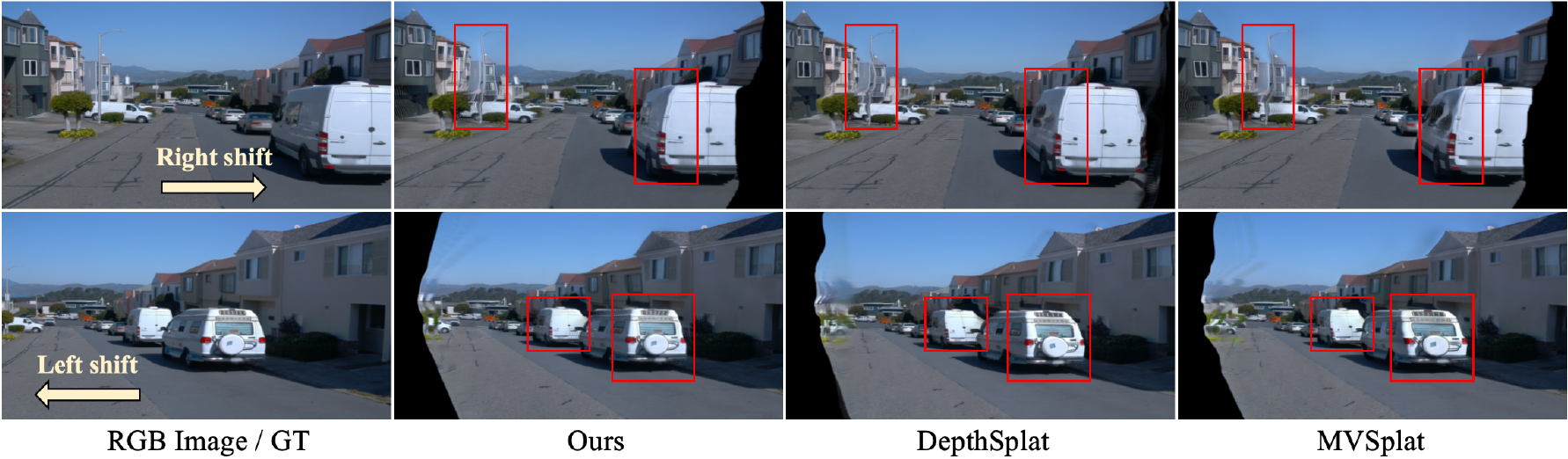}
   \vspace{-6mm}
   \caption{\textbf{Visual comparisons on view shifting.} The figure displays the performance of right and left shifting of the given images on the Waymo dataset. The challenging areas are marked with red rectangles. As observed, our model exhibits superior robustness under large viewpoint changes. }
   \vspace{-2mm}
   \label{vis_shift}
\end{figure*}

\subsection{Analyses on Multi-modal Inputs}
To ensure a fair comparison, we further constructed baseline networks with identical multi-modal inputs using state-of-the-art depth completion methods. Specifically, we re-implemented CFormer \cite{zhang2023completionformer} and BPNet \cite{tang2024bilateral} as the comparison targets, which take both image data and sparse depth as inputs and predict Gaussian parameters using multi-modal fused features. Also, we initialized these models with weights pre-trained on the KITTI depth completion dataset. By maintaining identical input modalities and training conditions, these baselines allow us to evaluate the performance gains attributable to the inclusion of depth data, independent of the architectural advancements in our framework.

 


\paragraph{Comparison Results}
As shown in Table \ref{ablation2}, it is evident that the inclusion of additional depth input alone does not significantly enhance the quality of novel view rendering. This demonstrates the limitations of relying solely on accurate depth prediction for high-quality rendering, further highlighting the critical need for our joint optimization. Additionally, as demonstrated in Fig. \ref{vis_depth}, our model effectively preserves fine structural details (e.g., car contours and pole geometries) without pre-training on depth completion tasks.

\paragraph{Robustness to depth quality}

We evaluate our model's robustness to depth quality degradation by randomly discarding approximately 50\% of LiDAR points during testing. As presented in the right part of Table \ref{ablation2}, ADGaussian remains competitive with full-depth-input performance under such conditions. The performance gap compared to baselines validates our cross-modal optimization framework's ability to compensate for missing depth information through effective appearance-geometry integration.

\subsection{Application: Novel-view Shifting}

The concept of novel-view shifting involves generating images from significantly varied perspectives compared to the original viewpoints present in the training data. This task is particularly demanding as it usually necessitates reliable depth estimations to handle substantial changes in viewpoint and scale. 
In this study, we further investigate the model's robustness in view shifting. Firstly, the Ground Truth right camera images provided in the KITTI dataset are used to evaluate the quantitative performance of view shifting. As depicted in Table \ref{shift}, our model significantly outperforms both MVSplat and DepthSplat in zero-shot view shifting from left to right cameras. It is noteworthy that our zero-shot view shifting results are only slightly lower than our normally trained model (PSNR: 23.60, SSIM: 0.776, LPIPS: 0.164). 
Moreover, visual comparisons on the Waymo dataset are presented in Fig. \ref{vis_shift}. ADGaussian exhibits exceptional view shifting quality, accurately maintaining object shapes and preserving intricate details even in texture-less regions during both leftward and rightward viewpoint changes. 

\begin{table}[t]
\caption{\textbf{Robust analyses on novel-view shifting on the KITTI dataset.} Models trained on multi-frame images are utilized directly to evaluate their capability for novel-view shifting, transitioning from the left camera to the right camera, all without additional fine-tuning.}
\vspace{-1mm}
\centering
\resizebox{0.77\columnwidth}{!}{
\begin{tabular}{lccc}
\toprule
Method & PSNR\(\uparrow\) & SSIM\(\uparrow\) & LPIPS\(\downarrow\) \\
\midrule
MVSplat \cite{chen2024mvsplat} & 14.39    &  0.474           &  0.382     \\
DepthSplat \cite{xu2024depthsplat} &   15.07       &  0.452           &    0.377    \\
Ours &  21.81       & 0.770            &    0.184   \\
\bottomrule
\end{tabular}}
\vspace{-2mm}
\label{shift}
\end{table}

\section{Conclusion}


This paper presents ADGaussian, a novel multimodal framework that advances generalizable street scene reconstruction via synergistic learning of visual and geometric features. We validate that jointly optimizing RGB imagery and sparse depth inputs significantly improves both geometric quality and visual fidelity. Comprehensive experiments on Waymo and KITTI datasets establish state-of-the-art performance in complex urban scenarios. The framework also shows strong zero-shot generalization for large viewpoint shifts while maintaining metric scale consistency. Current limitations include the sparsity of single-view information and the challenge of modeling dynamic objects. Extending the approach to multi-frame fusion could help address these gaps.

\bibliographystyle{IEEEtran}
\bibliography{ref}

@String(AAAI = {AAAI})

@article{kerbl20233d,
  title={3d gaussian splatting for real-time radiance field rendering.},
  author={Kerbl, Bernhard and Kopanas, Georgios and Leimk{\"u}hler, Thomas and Drettakis, George},
  journal={ACM Trans. Graph.},
  volume={42},
  number={4},
  pages={139--1},
  year={2023}
}

@article{yan2023street,
  title={Street Gaussians for Modeling Dynamic Urban Scenes.(2023)},
  author={Yan, Yunzhi and Lin, Haotong and Zhou, Chenxu and Wang, Weijie and Sun, Haiyang and Zhan, Kun and Lang, Xianpeng and Zhou, Xiaowei and Peng, Sida},
  year={2023}
}

@inproceedings{chen2024mvsplat,
  title={Mvsplat: Efficient 3d gaussian splatting from sparse multi-view images},
  author={Chen, Yuedong and Xu, Haofei and Zheng, Chuanxia and Zhuang, Bohan and Pollefeys, Marc and Geiger, Andreas and Cham, Tat-Jen and Cai, Jianfei},
  booktitle={European Conference on Computer Vision},
  pages={370--386},
  year={2024},
  organization={Springer}
}

@inproceedings{charatan2024pixelsplat,
  title={pixelsplat: 3d gaussian splats from image pairs for scalable generalizable 3d reconstruction},
  author={Charatan, David and Li, Sizhe Lester and Tagliasacchi, Andrea and Sitzmann, Vincent},
  booktitle={Proceedings of the IEEE/CVF Conference on Computer Vision and Pattern Recognition},
  pages={19457--19467},
  year={2024}
}

@inproceedings{li2024ggrt,
  title={Ggrt: Towards pose-free generalizable 3d gaussian splatting in real-time},
  author={Li, Hao and Gao, Yuanyuan and Wu, Chenming and Zhang, Dingwen and Dai, Yalun and Zhao, Chen and Feng, Haocheng and Ding, Errui and Wang, Jingdong and Han, Junwei},
  booktitle={European Conference on Computer Vision},
  pages={325--341},
  year={2024},
  organization={Springer}
}

@article{han2024ggs,
  title={GGS: Generalizable Gaussian Splatting for Lane Switching in Autonomous Driving},
  author={Han, Huasong and Zhou, Kaixuan and Long, Xiaoxiao and Wang, Yusen and Xiao, Chunxia},
  journal={arXiv preprint arXiv:2409.02382},
  year={2024}
}

@article{xu2024depthsplat,
  title={Depthsplat: Connecting gaussian splatting and depth},
  author={Xu, Haofei and Peng, Songyou and Wang, Fangjinhua and Blum, Hermann and Barath, Daniel and Geiger, Andreas and Pollefeys, Marc},
  journal={arXiv preprint arXiv:2410.13862},
  year={2024}
}

@article{yang2024depth,
  title={Depth Anything V2},
  author={Yang, Lihe and Kang, Bingyi and Huang, Zilong and Zhao, Zhen and Xu, Xiaogang and Feng, Jiashi and Zhao, Hengshuang},
  journal={arXiv preprint arXiv:2406.09414},
  year={2024}
}

@inproceedings{ranftl2021vision,
  title={Vision transformers for dense prediction},
  author={Ranftl, Ren{\'e} and Bochkovskiy, Alexey and Koltun, Vladlen},
  booktitle={Proceedings of the IEEE/CVF international conference on computer vision},
  pages={12179--12188},
  year={2021}
}

@inproceedings{sun2020scalability,
  title={Scalability in perception for autonomous driving: Waymo open dataset},
  author={Sun, Pei and Kretzschmar, Henrik and Dotiwalla, Xerxes and Chouard, Aurelien and Patnaik, Vijaysai and Tsui, Paul and Guo, James and Zhou, Yin and Chai, Yuning and Caine, Benjamin and others},
  booktitle={Proceedings of the IEEE/CVF conference on computer vision and pattern recognition},
  pages={2446--2454},
  year={2020}
}

@inproceedings{geiger2012we,
  title={Are we ready for autonomous driving? the kitti vision benchmark suite},
  author={Geiger, Andreas and Lenz, Philip and Urtasun, Raquel},
  booktitle={2012 IEEE conference on computer vision and pattern recognition},
  pages={3354--3361},
  year={2012},
  organization={IEEE}
}

@article{wang2004image,
  title={Image quality assessment: from error visibility to structural similarity},
  author={Wang, Zhou and Bovik, Alan C and Sheikh, Hamid R and Simoncelli, Eero P},
  journal={IEEE transactions on image processing},
  volume={13},
  number={4},
  pages={600--612},
  year={2004},
  publisher={IEEE}
}

@inproceedings{zhang2018unreasonable,
  title={The unreasonable effectiveness of deep features as a perceptual metric},
  author={Zhang, Richard and Isola, Phillip and Efros, Alexei A and Shechtman, Eli and Wang, Oliver},
  booktitle={Proceedings of the IEEE conference on computer vision and pattern recognition},
  pages={586--595},
  year={2018}
}

@inproceedings{tang2024bilateral,
  title={Bilateral Propagation Network for Depth Completion},
  author={Tang, Jie and Tian, Fei-Peng and An, Boshi and Li, Jian and Tan, Ping},
  booktitle={Proceedings of the IEEE/CVF Conference on Computer Vision and Pattern Recognition},
  pages={9763--9772},
  year={2024}
}

@inproceedings{liu2024mvsgaussian,
  title={Mvsgaussian: Fast generalizable gaussian splatting reconstruction from multi-view stereo},
  author={Liu, Tianqi and Wang, Guangcong and Hu, Shoukang and Shen, Liao and Ye, Xinyi and Zang, Yuhang and Cao, Zhiguo and Li, Wei and Liu, Ziwei},
  booktitle={European Conference on Computer Vision},
  pages={37--53},
  year={2024},
  organization={Springer}
}

@article{smart2024splatt3r,
  title={Splatt3r: Zero-shot gaussian splatting from uncalibrated image pairs},
  author={Smart, Brandon and Zheng, Chuanxia and Laina, Iro and Prisacariu, Victor Adrian},
  journal={arXiv preprint arXiv:2408.13912},
  year={2024}
}

@article{ye2024no,
  title={No pose, no problem: Surprisingly simple 3d gaussian splats from sparse unposed images},
  author={Ye, Botao and Liu, Sifei and Xu, Haofei and Li, Xueting and Pollefeys, Marc and Yang, Ming-Hsuan and Peng, Songyou},
  journal={arXiv preprint arXiv:2410.24207},
  year={2024}
}

@inproceedings{zheng2024gps,
  title={Gps-gaussian: Generalizable pixel-wise 3d gaussian splatting for real-time human novel view synthesis},
  author={Zheng, Shunyuan and Zhou, Boyao and Shao, Ruizhi and Liu, Boning and Zhang, Shengping and Nie, Liqiang and Liu, Yebin},
  booktitle={Proceedings of the IEEE/CVF Conference on Computer Vision and Pattern Recognition},
  pages={19680--19690},
  year={2024}
}

@article{mildenhall2021nerf,
  title={Nerf: Representing scenes as neural radiance fields for view synthesis},
  author={Mildenhall, Ben and Srinivasan, Pratul P and Tancik, Matthew and Barron, Jonathan T and Ramamoorthi, Ravi and Ng, Ren},
  journal={Communications of the ACM},
  volume={65},
  number={1},
  pages={99--106},
  year={2021},
  publisher={ACM New York, NY, USA}
}

@article{song2024divide,
  title={Divide and conquer: Improving multi-camera 3D perception with 2D semantic-depth priors and input-dependent queries},
  author={Song, Qi and Hu, Qingyong and Zhang, Chi and Chen, Yongquan and Huang, Rui},
  journal={IEEE Transactions on Image Processing},
  volume={33},
  pages={897--909},
  year={2024},
  publisher={IEEE}
}

@inproceedings{yang2023reconstructing,
  title={Reconstructing objects in-the-wild for realistic sensor simulation},
  author={Yang, Ze and Manivasagam, Sivabalan and Chen, Yun and Wang, Jingkang and Hu, Rui and Urtasun, Raquel},
  booktitle={2023 IEEE International Conference on Robotics and Automation (ICRA)},
  pages={11661--11668},
  year={2023},
  organization={IEEE}
}

@article{wang2023cadsim,
  title={Cadsim: Robust and scalable in-the-wild 3d reconstruction for controllable sensor simulation},
  author={Wang, Jingkang and Manivasagam, Sivabalan and Chen, Yun and Yang, Ze and B{\^a}rsan, Ioan Andrei and Yang, Anqi Joyce and Ma, Wei-Chiu and Urtasun, Raquel},
  journal={arXiv preprint arXiv:2311.01447},
  year={2023}
}

@article{lu2024drivingrecon,
  title={DrivingRecon: Large 4D Gaussian Reconstruction Model For Autonomous Driving},
  author={Lu, Hao and Xu, Tianshuo and Zheng, Wenzhao and Zhang, Yunpeng and Zhan, Wei and Du, Dalong and Tomizuka, Masayoshi and Keutzer, Kurt and Chen, Yingcong},
  journal={arXiv preprint arXiv:2412.09043},
  year={}
}

@inproceedings{szymanowicz2024splatter,
  title={Splatter image: Ultra-fast single-view 3d reconstruction},
  author={Szymanowicz, Stanislaw and Rupprecht, Chrisitian and Vedaldi, Andrea},
  booktitle={Proceedings of the IEEE/CVF Conference on Computer Vision and Pattern Recognition},
  pages={10208--10217},
  year={2024}
}

@article{wang2024freesplat,
  title={FreeSplat: Generalizable 3D Gaussian Splatting Towards Free-View Synthesis of Indoor Scenes},
  author={Wang, Yunsong and Huang, Tianxin and Chen, Hanlin and Lee, Gim Hee},
  journal={arXiv preprint arXiv:2405.17958},
  year={2024}
}

@inproceedings{zhao2024tclc,
  title={TCLC-GS: Tightly Coupled LiDAR-Camera Gaussian Splatting for Autonomous Driving: Supplementary Materials},
  author={Zhao, Cheng and Sun, Su and Wang, Ruoyu and Guo, Yuliang and Wan, Jun-Jun and Huang, Zhou and Huang, Xinyu and Chen, Yingjie Victor and Ren, Liu},
  booktitle={European Conference on Computer Vision},
  pages={91--106},
  year={2024},
  organization={Springer}
}

@inproceedings{chung2024depth,
  title={Depth-regularized optimization for 3d gaussian splatting in few-shot images},
  author={Chung, Jaeyoung and Oh, Jeongtaek and Lee, Kyoung Mu},
  booktitle={Proceedings of the IEEE/CVF Conference on Computer Vision and Pattern Recognition},
  pages={811--820},
  year={2024}
}

@article{turkulainen2024dn,
  title={DN-Splatter: Depth and Normal Priors for Gaussian Splatting and Meshing},
  author={Turkulainen, Matias and Ren, Xuqian and Melekhov, Iaroslav and Seiskari, Otto and Rahtu, Esa and Kannala, Juho},
  journal={arXiv preprint arXiv:2403.17822},
  year={2024}
}

@inproceedings{piccinelli2024unidepth,
  title={UniDepth: Universal Monocular Metric Depth Estimation},
  author={Piccinelli, Luigi and Yang, Yung-Hsu and Sakaridis, Christos and Segu, Mattia and Li, Siyuan and Van Gool, Luc and Yu, Fisher},
  booktitle={Proceedings of the IEEE/CVF Conference on Computer Vision and Pattern Recognition},
  pages={10106--10116},
  year={2024}
}

@inproceedings{wewer2024latentsplat,
  title={latentsplat: Autoencoding variational gaussians for fast generalizable 3d reconstruction},
  author={Wewer, Christopher and Raj, Kevin and Ilg, Eddy and Schiele, Bernt and Lenssen, Jan Eric},
  booktitle={European Conference on Computer Vision},
  pages={456--473},
  year={2024},
  organization={Springer}
}

@article{kung2024lihi,
  title={LiHi-GS: LiDAR-Supervised Gaussian Splatting for Highway Driving Scene Reconstruction},
  author={Kung, Pou-Chun and Zhang, Xianling and Skinner, Katherine A and Jaipuria, Nikita},
  journal={arXiv preprint arXiv:2412.15447},
  year={2024}
}

@article{khan2024autosplat,
  title={Autosplat: Constrained gaussian splatting for autonomous driving scene reconstruction},
  author={Khan, Mustafa and Fazlali, Hamidreza and Sharma, Dhruv and Cao, Tongtong and Bai, Dongfeng and Ren, Yuan and Liu, Bingbing},
  journal={arXiv preprint arXiv:2407.02598},
  year={2024}
}

@inproceedings{zhang2023completionformer,
  title={Completionformer: Depth completion with convolutions and vision transformers},
  author={Zhang, Youmin and Guo, Xianda and Poggi, Matteo and Zhu, Zheng and Huang, Guan and Mattoccia, Stefano},
  booktitle={Proceedings of the IEEE/CVF conference on computer vision and pattern recognition},
  pages={18527--18536},
  year={2023}
}

@inproceedings{wang2024dust3r,
  title={Dust3r: Geometric 3d vision made easy},
  author={Wang, Shuzhe and Leroy, Vincent and Cabon, Yohann and Chidlovskii, Boris and Revaud, Jerome},
  booktitle={Proceedings of the IEEE/CVF Conference on Computer Vision and Pattern Recognition},
  pages={20697--20709},
  year={2024}
}

@inproceedings{leroy2024grounding,
  title={Grounding image matching in 3d with mast3r},
  author={Leroy, Vincent and Cabon, Yohann and Revaud, J{\'e}r{\^o}me},
  booktitle={European Conference on Computer Vision},
  pages={71--91},
  year={2024},
  organization={Springer}
}

@inproceedings{chen2024g3r,
  title={G3r: Gradient guided generalizable reconstruction},
  author={Chen, Yun and Wang, Jingkang and Yang, Ze and Manivasagam, Sivabalan and Urtasun, Raquel},
  booktitle={European Conference on Computer Vision},
  pages={305--323},
  year={2024},
  organization={Springer}
}

@inproceedings{tian2025drivingforward,
  title={Drivingforward: Feed-forward 3d gaussian splatting for driving scene reconstruction from flexible surround-view input},
  author={Tian, Qijian and Tan, Xin and Xie, Yuan and Ma, Lizhuang},
  booktitle={Proceedings of the AAAI Conference on Artificial Intelligence},
  volume={39},
  number={7},
  pages={7374--7382},
  year={2025}
}

@inproceedings{yan2025streetcrafter,
  title={Streetcrafter: Street view synthesis with controllable video diffusion models},
  author={Yan, Yunzhi and Xu, Zhen and Lin, Haotong and Jin, Haian and Guo, Haoyu and Wang, Yida and Zhan, Kun and Lang, Xianpeng and Bao, Hujun and Zhou, Xiaowei and others},
  booktitle={Proceedings of the Computer Vision and Pattern Recognition Conference},
  pages={822--832},
  year={2025}
}

@article{zhao2025recondreamer++,
  title={Recondreamer++: Harmonizing generative and reconstructive models for driving scene representation},
  author={Zhao, Guosheng and Wang, Xiaofeng and Ni, Chaojun and Zhu, Zheng and Qin, Wenkang and Huang, Guan and Wang, Xingang},
  journal={arXiv preprint arXiv:2503.18438},
  year={}
}

@article{jiang2024li,
  title={Li-gs: Gaussian splatting with lidar incorporated for accurate large-scale reconstruction},
  author={Jiang, Changjian and Gao, Ruilan and Shao, Kele and Wang, Yue and Xiong, Rong and Zhang, Yu},
  journal={IEEE Robotics and Automation Letters},
  year={2024},
  publisher={IEEE}
}

@article{huang2024textit,
  title={S3Gaussian: Self-Supervised Street Gaussians for Autonomous Driving},
  author={Huang, Nan and Wei, Xiaobao and Zheng, Wenzhao and An, Pengju and Lu, Ming and Zhan, Wei and Tomizuka, Masayoshi and Keutzer, Kurt and Zhang, Shanghang},
  journal={arXiv preprint arXiv:2405.20323},
  year={2024}
}

\end{document}